# Zero-Shot Document-Level Biomedical Relation Extraction via Scenario-based Prompt Design in Two-Stage with LLM

Lei Zhao[1], Ling Kang[1], Quan Guo*[1]

*1 Neusoft Research Institution, Dalian Neusoft University of Information, Dalian, China*

* Corresponding author. Email: guoquan@neusoft.edu.cn

**Abstract**

With the advent of artificial intelligence (AI), many researchers are attempting to extract structured information from document-level biomedical literature by fine-tuning large language models (LLMs). However, they face significant challenges such as the need for expensive hardware, like high-performance GPUs and the high labor costs associated with annotating training datasets, especially in biomedical realm. Recent research on LLMs, such as GPT-4 and Llama3, has shown promising performance in zero-shot settings, inspiring us to explore a novel approach to achieve the same results from unannotated full documents using general LLMs with lower hardware and labor costs. Our approach combines two major stages: named entity recognition (NER) and relation extraction (RE). NER identifies chemical, disease and gene entities from the document with synonym and hypernym extraction using an LLM with a crafted prompt. RE extracts relations between entities based on predefined relation schemas and prompts. To enhance the effectiveness of prompt, we propose a five-part template structure and a scenario-based prompt design principles, along with evaluation method to systematically assess the prompts. Finally, we evaluated our approach against fine-tuning and pre-trained models on two biomedical datasets: ChemDisGene and CDR. The experimental results indicate that our proposed method can achieve comparable accuracy levels to fine-tuning and pre-trained models but with reduced human and hardware expenses.

***Keywords:*** *Two-stage Method, Scenario-based Prompt Design Principles, Five-part Prompt Template Structure, Zero Shot, Document-level Biomedical Relation Extraction, Large Language Model*

## Introduction

Artificial intelligence (AI) plays more critical roles for extracting biomedical structured information from unstructured literatures, e.g. relations for drug-drug interaction(Datta et al., 2022; Sahu and Anand, 2017), drug-disease interaction(Mongia et al., 2022; Wang et al., 2017), food-disease interaction(Cenikj et al., 2023; Zhao et al., 2024), chemical reactions(Vangala et al., 2024), etc. The majority AI models rely on high-quality human annotated datasets used for training. The commercial dataset is limited by subscription fees or institutional access, while open-source dataset is inefficient application scenarios. Though utilizing deep learning techniques, researchers have developed Large Language Models (LLM) (Ai et al., 2024; Dagdelen et al., 2024) supported less dataset to fine-tuning the parameters to improve the accuracy, fine-tuning relies on high expensive hardware equipment e.g. GPU. Recent research on LLMs, such as GPT-4 released, has shown promising performance in zero-shot settings (zero-shot implies that the model can extract structured information from unstructured text without being explicitly trained on that task for each new class or type of information), some researchers is attempting to propose prompt based zero shot approaches for the same tasks in general scenarios(Xue et al., 2024, n.d.), it's inspiring us to explore a novel approach extract biomedical information relations from document level text.

The relation extracts have two major tasks, one is named naming recognition (NER), recognize the named entity from unstructured text, the other is relation extraction (RE), extract the relations between entities. To complete the task, the supervised models need training annotated dataset, then predict fresh data:

TTM-RE(Gao et al., 2024) is a supervised model that introduces a novel approach for document-level relation extraction. TTM-RE integrates a trainable memory module, known as the Token Turing Machine (TTM), with a noise-robust loss function designed for positive-unlabeled settings. The TTM-RE model is designed to effectively leverage large-scale, noisy training data to improve the performance of relation extraction tasks. The model is evaluated on ChemDisGene(Zhang et al., 2022) dataset in the biomedical domain. The limitation mentioned in this paper: the performance it attained is still limited compared to supervised methods on the same task. Relation prediction still requires a large amount of data.

A Multi-view Merge Representation model (MMR)(Zhang et al., 2024) is also a supervised model based on BioBERT(Lee et al., 2020) that has achieved excellent results in document-level relation extraction tasks. This is a method that capture entity and entity-pair representation of the document. It utilizes prior knowledge and a pre-trained transformer encoder to capture entity semantic representation. Then it employs the U-Net layer and Graph Convolution Network layer to capture global entity-pair representation. Finally, it gets a specific merged representation for each entity pair to be classified. The paper mentioned that MMR relies on prior knowledge from the CTD, which may lead to deficiencies when confronted with novel knowledge.

Via fast growing computing power of hardware, LLM such as ChatGPT has more supper parameters enabled remembering more expertise entities with more reasoning capability. This development has transformed the traditional training-predict into zero-shot:

ChatIE zero-shot information extract(IE)(Wei et al., 2024) aims to build IE systems from the unannotated text. ChatIE transforms IE task into a multi-turn question-answering problem with a two-stage framework. With the power of ChatGPT, it extensively evaluates its framework on three IE tasks: entity-relation triple extract, NER, and event extraction. However, ChatIE prompt might potentially miss certain entities or even make incorrect judgments due to the complexity of document structures.

This paper aims to explore a novel approach for zero-shot document-level biomedical relation extraction, building upon the innovative framework established by ChatIE. We propose a two-stage framework that leverages the strengths of LLM while mitigating their inherent limitations. To construct effective prompts for each stage, we propose a scenario-based prompt design principles and utilize a five-part template structure for prompt design, which prevent potential false negative and false positive errors that may be caused by LLM. Finally, we conducted comprehensive experiments across ChemDisGene and CDR. The empirical results are promising, demonstrating performance that is not only impressive but also comparable to that of supervised models.

## Materials and Methods

### *Two-stage framework*

The two-stage framework is illustrated in Fig. 1 and comprises the following steps: Stage I: Prepare Prompt Templates: designed prompt templates for NER task. Pre-process Dataset: The original dataset was transformed into a unified format. Process NER: Identified entities and their attributes and established synonym and hypernym relations. Stage II: Prepare Prompt Templates: further designed the prompt templates for RE task. Define RE Schema: define a triple element relation schema. Process RE: Utilizing the defined schema, extracted relations between entities and generated the final structured dataset. Evaluate: assess the performance of both NER and RE using precision, recall, and F1 score.

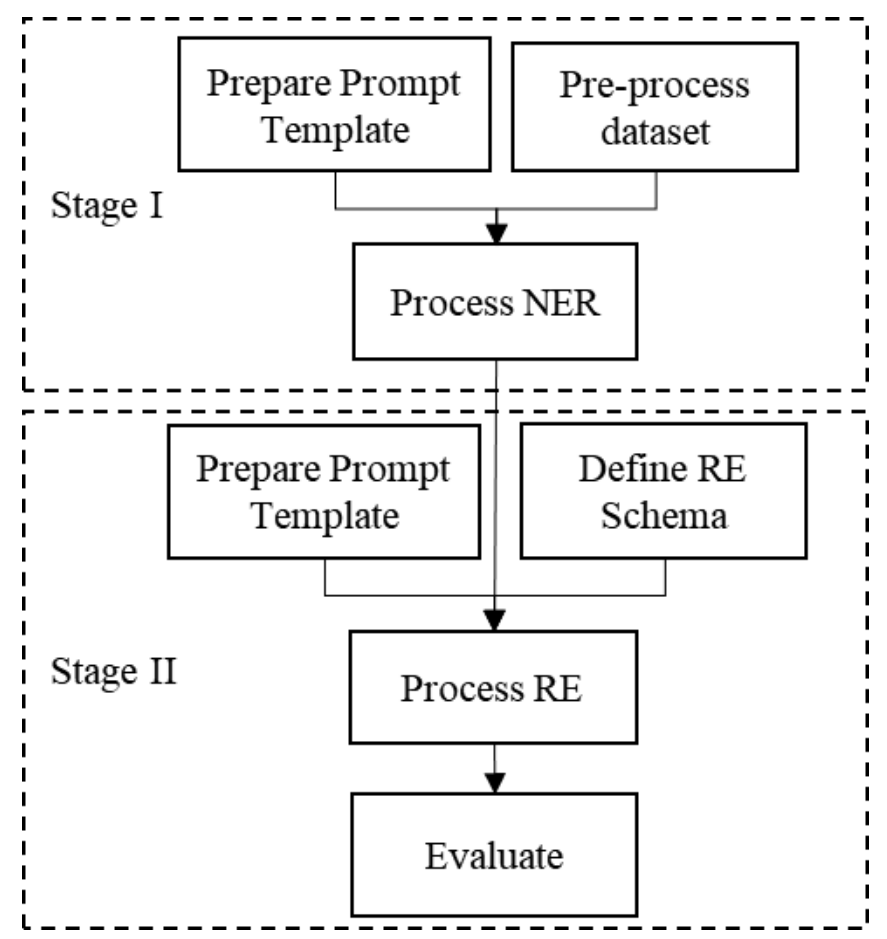

**Fig. 1** Two-stage framework with seven steps

**Prepare prompt template for Stage I**

Prompt template $\boldsymbol{PT_{NER}}$ in Stage I is structured with five-part template structure as depicted in Fig. 2. Context contains document ***D***. Requirement defined the specific request to ***LLM***. Positive Scenarios cover all possible examples. Negative scenarios offer exclusion examples. Output Format defines data format for subsequent processing.

*Relative efficacy and **toxicity** of **netilmicin** and **tobramycin** in oncology patients. We prospectively compared the efficacy and safety of **netilmicin sulfate** or **tobramycin sulfate** in conjunction with **piperacillin sodium** in 118 immunocompromised patients with presumed severe **infections**. The two treatment regimens were equally efficacious. **Nephrotoxicity** occurred in a similar proportion in patients treated with **netilmicin** and **tobramycin** (17% vs 11%). **Ototoxicity** occurred in four (9.5%) of 42 **netilmicin** and **piperacillin** and in 12 (22%) of 54 **tobramycin** and **piperacillin**-treated patients. Of those evaluated with posttherapy audiograms, three of four netilmicin and piperacillin-treated patients had auditory thresholds return to baseline compared with one of nine **tobramycin** and **piperacillin**-treated patients. The number of greater than or equal to 15-dB increases in auditory threshold as a proportion of total greater than or equal to 15-dB changes (increases and decreases) was significantly lower in **netilmicin** and **piperacillin**- vs **tobramycin** and **piperacillin**-treated patients (18 of 78 vs 67 of 115). We conclude that aminoglycoside-associated **ototoxicity** was less severe and more often reversible with **netilmicin** than with **tobramycin**.*

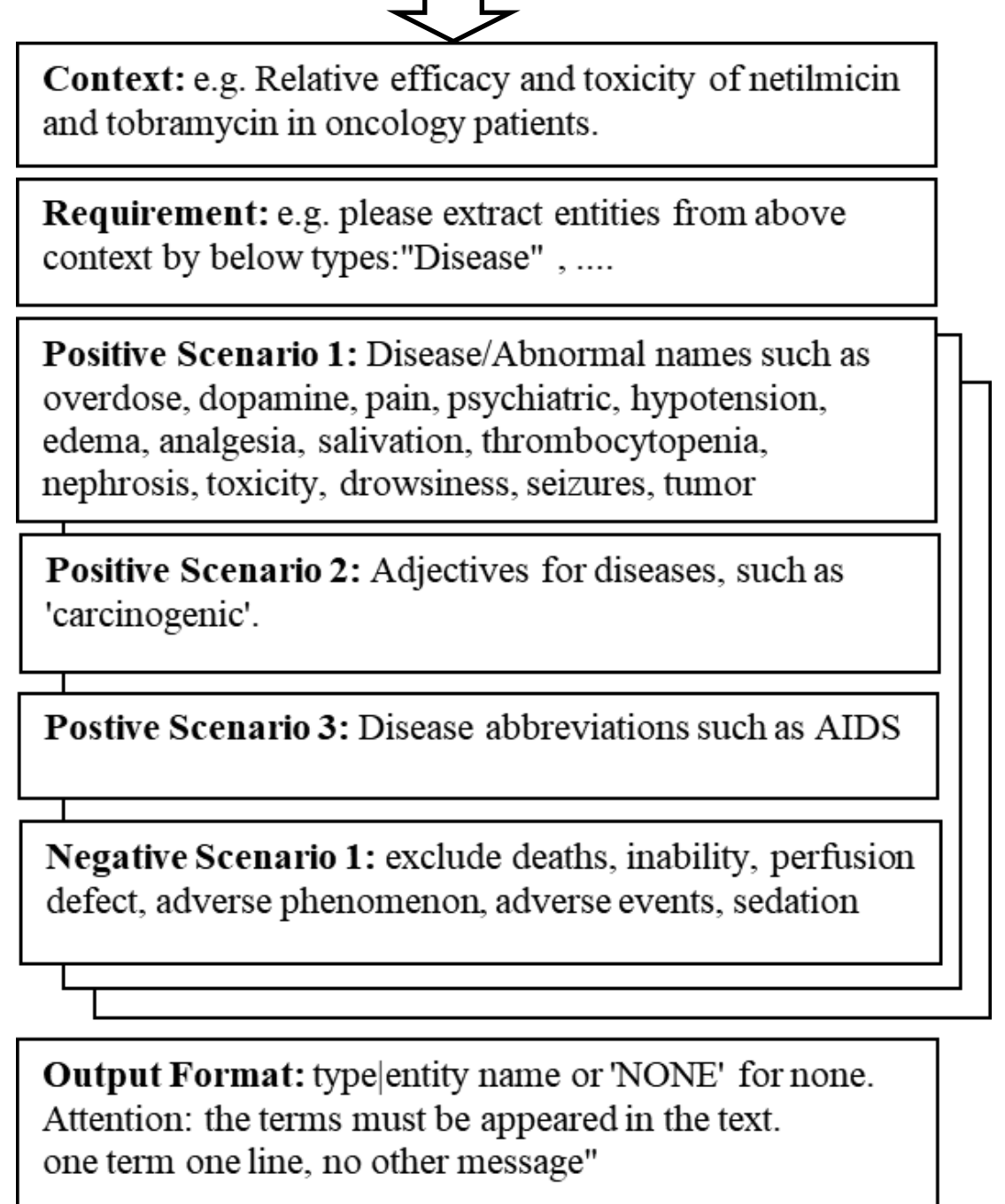

**Fig.2** NER prompt template structure: an example for disease entity recognition on CDR. Green fonts denote chemical entities and red fonts denote disease entities.

Drawing inspiration from CR(Xiao et al., 2022), we extract synonym and hypernym relations between entities during this phase. $\boldsymbol{PT_{synonym}}$ is designed to identify synonym relation between entities, while $\boldsymbol{PT_{hypernym}}$ is intended for hypernym relations.

**Pre-process dataset**

The dataset consists of documents $D=\{D_1, D_2, D_3, \ldots, D_n\ldots, D_N\}$, each document $\boldsymbol{D_n}$ is composed of sentences $D_n=\{S_1,S_2,S_3,\ldots,S_m,\ldots,S_M\}$, and each sentence $\boldsymbol{S_m}$ consists of words $S_m=\{W_1,W_2,W_3,\ldots,W_p,\ldots,W_P\}$.

**Process NER**

Input the document $\boldsymbol{D}$ and entity labels $\boldsymbol{L}$ along with the prompt template $\boldsymbol{PT_{NER}}$ into $\boldsymbol{LLM}$ to obtain recognized entities $\boldsymbol{N}$:

$$N = LLM(S, L, PT_{NER}) \tag{1}$$

where $\boldsymbol{N}=\{e, l, index_{\text{sentence}}, [index_{\text{start}}, index_{\text{end}}]_{\text{position}}, index_{\text{document}}\}$, $e\in\boldsymbol{W}$, $l\in\boldsymbol{L}$. $index_{\text{sentence}}$ indicates which sentence the entity appears in $\boldsymbol{D}_n$, $[index_{\text{start}}, index_{\text{end}}]_{\text{position}}$ indicates the start position and end position of the entity within the sentence.

Pair up each entity $e$, resulting in $\boldsymbol{P}=\{(e_1, e_2), (e_1, e_3), (e_2, e_3) \ldots, (e_n, e_m)\}$. Input $\boldsymbol{P}$, $\boldsymbol{D}$ and $\boldsymbol{PT_{synonym}}$ into $\boldsymbol{LLM}$, $\boldsymbol{N_{synonym}}$ is obtained as follow:

$$N_{\boldsymbol{synonym}} = LLM(D, P, PT_{\boldsymbol{synonym}}) \tag{2}$$

where $\boldsymbol{N_{synonym}}=\{N, index_{\boldsymbol{global}}\}$, where synonyms or same words share the same global index $index_{\boldsymbol{global}}$.

For example, as shown in Table 1, the entity named *netilmicin sulfate* is treated as the synonym of the entity named *netilmicin* in this document.

Pair up $e$ with unique $index_{\boldsymbol{synonym}}$ to obtain $\boldsymbol{P'}=\{(e_{1'}, e_{2'}), (e_{1'}, e_{3'}), (e_{2'}, e_{3'}) \ldots, (e_{n'}, e_{m'})\}$. Input $\boldsymbol{P'}$, $\boldsymbol{D}$ and $\boldsymbol{PT_{hypernym}}$ into $\boldsymbol{LLM}$, hypernym result $\boldsymbol{N_{predicted}}$ as follow:

$$N_{\boldsymbol{predicted}} = LLM(D, P', PT_{\boldsymbol{hypernym}}) \tag{3}$$

where $\boldsymbol{N_{predicted}}=\{\boldsymbol{N_{synonym}}, index_{\boldsymbol{hypernym}}\}$, where $index_{\boldsymbol{hypernym}}$ is either $index_{\boldsymbol{global}}$ of the hypernym entity or -1.

For example, as shown in Table 1, *Toxicity* is the hypernym of *ototoxicity*. Consequently, the hypernym index for *ototoxicity* corresponds to the $index_{\text{global}}$ (global index) of *Toxicity*, which is 0.

**Table 1** Partial Sample of NER Predicted Results

| Entity | Label | Sent. Index | Pos. Index | Global Index | Hypernym Index | Document Index |
|---|---|---|---|---|---|---|
| toxicity | Disease | 0 | [3,4] | 0 | -1 | 3535719 |
| netilmicin | Chemical | 0 | [5,6] | 1 | -1 | 3535719 |
| tobramycin | Chemical | 0 | [7,8] | 2 | -1 | 3535719 |
| netilmicin sulfate | Chemical | 1 | [8,10] | 1 | -1 | 3535719 |
| ototoxicity | Disease | 4 | [0,1] | 3 | 0 | 3535719 |
| … | … | … | … | … | … | … |

**Prepare prompt template for Stage II**

Prompt template $\boldsymbol{PT_{RE}}$ in Stage II, as depicted as Fig 3, is structured with five parts: Context includes document $\boldsymbol{D}$. Requirement defined the specific request to $\boldsymbol{LLM}$. Positive scenario encompasses all possible cases $\boldsymbol{A}$, each positive answer $\boldsymbol{a}$ is linked with a relation $r$, $r\in\boldsymbol{R}$, where $\boldsymbol{R}$ is relation set. Negative scenario covers $\boldsymbol{LLM}$ false responses. Output Format defines output data format for subsequent processing.

**Context:** e.g. Relative efficacy and toxicity of netilmicin and tobramycin in oncology patients. …….

**Requirement:** You are biomedical expert, refer to above biomedical article, please give answer '$' or '~' only based on below conditions:
Assumptions:
…….

**Positive Scenario 1:** if the article gave the conclusion that the interaction between {head} and another chemical induce {tail}, or {head} induce {tail}, answer '$'

**Positive Scenario 2:** else if the article mentioned {tail} was reported in the percentage of patients because of {head}, answer '$'

**Negative Scenario 1:** else the article mentioned '{head} is mediating/attenuating {tail}, answer '~'.

**Negative Scenario 2:** else if {head} and {tail} have no relationship, answer '~'.

**Output Format:** please give answer '$' or '~'
**DO NOT output any other message**

**Fig. 3** RE prompt template structure: an example of chemical-induce-disease

**Define RE Schema**

A predefined schema set $\boldsymbol{RS}$ is established as $rs=\{l_{\text{head}}, r, l_{\text{tail}}\}$ where $l_{\text{head}}, l_{\text{tail}} \in \boldsymbol{L}$, $rs \in \boldsymbol{RS}$, $r \in \boldsymbol{R}$.

**Process RE**

The overall procedure of this phase is presented in Algorithm 1. To extract relation $\boldsymbol{R}$ from $\boldsymbol{D}$ to obtain final RE result $\boldsymbol{RE}_{\textbf{predicted}}$. Entity Filtering: $\boldsymbol{E}_{\textbf{head}}$ is obtained by filtering $\boldsymbol{N}_{\textbf{predicted}}$ for entities where the label $l$ is $l_{\text{head}}$ and not a hypernym and then taking the unique entities based on $index_{\text{global}}$. $E_{\text{tail}}$ is identified similarly. Entity Pairing: entities from $E_{\text{head}}$ and $E_{\text{tail}}$ are paired to create a set of $\boldsymbol{P}$ where each pair is denoted as $(e_{\text{head}}, e_{\text{tail}})$. We performed hypernym filtering similar to Gu et al. (Gu et al., 2016). LLM Input and Response: $\boldsymbol{P}$ along with $\boldsymbol{PT_{RE}}$, $\boldsymbol{D}$ are input into $\boldsymbol{LLM}$ to receive a single answer $\boldsymbol{a}$. Confirmation and Addition: If $\boldsymbol{a}$ is not negative answer, confirming the relation $r$, the tuple $(e_{\text{head}}, r, e_{\text{tail}})$ is added into $\boldsymbol{RE}_{\textbf{predicted}}$. Synonym Inclusion: all synonyms of $e_{\text{head}}$ and $e_{\text{tail}}$ are considered to have the same relation, thus all $\boldsymbol{Synonyms}_{\textbf{head}}$ and $\boldsymbol{Synonyms}_{\textbf{tail}}$ are also included in $\boldsymbol{RE}_{\textbf{predicted}}$.

**Algorithm 1** Process RE

**Input:** **D**, **RS**, $\mathbf{N}_{\textbf{predicted}}$, $\mathbf{PT}_{\mathbf{RE}}$, **R**, **A**
**Output:** Final RE $\boldsymbol{RE}_{\textbf{predicted}}$
1: **for each** $\boldsymbol{r}$ **in** $\boldsymbol{R}$ **do**
2: $l_{\text{head}}, l_{\text{tail}} \leftarrow$ **select_schema(**$\boldsymbol{RS}$**,** $\boldsymbol{r}$**)** , $\boldsymbol{pt_r} \leftarrow$ **select_template(**$\boldsymbol{PT}_{\mathbf{RE}}$ **,**$\boldsymbol{r}$**)**
3: $Set_{\textbf{hypernym}} \leftarrow$ **select_column(**$\mathbf{index}_{\textbf{hypernym}}$**,** $\mathbf{N}_{\textbf{predicted}}$**)**
4: $E_{\textbf{head}} \leftarrow$ **unique(filter(**$N_{\textbf{predicted}}$**,** $l$ is $l_{\textbf{head}}$ and $index_{\textbf{global}}$ not exist in $Set_{\textbf{hypernym}}$**)** , $index_{\textbf{global}}$**)**
5: $E_{\textbf{tail}} \leftarrow$ **unique(filter(**$N_{\textbf{predicted}}$**,** $l$ is $l_{\textbf{tail}}$, and $index_{\textbf{global}}$ not exist in $Set_{\textbf{hypernym}}$**)**, $index_{\textbf{global}}$**)**
6: $P \leftarrow$ **pair_up(**$E_{\textbf{head}}$**,** $E_{\textbf{tail}}$**)**
7: **for** each $\boldsymbol{p}$ as $(e_{\text{head}}, e_{\text{tail}})$ in $\boldsymbol{P}$ **do**
8: $\boldsymbol{a} \leftarrow$ **LLM (**$\boldsymbol{pt_r}$**, p, D)**
9: **if** $a$ in $\boldsymbol{A}$ **then**
10: $\boldsymbol{RE}_{\textbf{predicted}}$.push($e_{\text{head}}$,r,$e_{\text{tail}}$)
11: $\boldsymbol{Synonyms}_{\textbf{head}} \leftarrow$ **find_ synonyms (**$\mathbf{e}_{\textbf{head}}$**,** $\boldsymbol{N}_{\textbf{predicted}}$**)**
12: $\boldsymbol{Synonyms}_{\textbf{tail}} \leftarrow$ **find_ synonyms (**$\mathbf{e}_{\textbf{tail}}$**,** $\boldsymbol{N}_{\textbf{predicted}}$**)**
13: $\boldsymbol{RE}_{\textbf{predicted}}$.push(each in $\boldsymbol{Synonyms}_{\textbf{head}}$ ,$\boldsymbol{r}$, each in $\boldsymbol{Synonyms}_{\textbf{tail}}$)
14: **end**

15:    **end**
16: **end**
17: **return** $\boldsymbol{RE}_{\mathbf{predicted}}$

The example of $\boldsymbol{RE}_{\mathbf{predicted}}$ is shown in Table 2.

**Table 2** Partial Sample of RE Predicted Result

| Head Entity | Label | Global Index | Relation | Tail Entity | Label | Global Index |
|---|---|---|---|---|---|---|
| netilmicin | Chemical | 1 | induced | ototoxicity | Disease | 3 |
| tobramycin | Chemical | 2 | induced | ototoxicity | Disease | 3 |
| netilmicin sulfate | Chemical | 1 | induced | ototoxicity | Disease | 3 |
| … | … | … | … | … | … | … |

**Evaluate**

We evaluate the performance of $\boldsymbol{N}_{\mathbf{predicted}}$ against the annotated dataset $\boldsymbol{N}_{\mathbf{annotated}}$ as well as $\boldsymbol{RE}_{\mathbf{predicted}}$ against $\boldsymbol{RE}_{\mathbf{annotated}}$, utilizing Precision, Recall and F1 scores.

### *Scenario-based prompt design principles*

In the five-part prompt template structure, we proposed positive scenario and negative scenario. Both are designed by our scenario-based prompt design principles. These principles are derived from our empirical insights from applying LLMs to biomedical literature processing.

Principle 1: *Single choice question:* Use single-choice questions rather than multiple-choice or free-answer questions.

Principle 2: *Numeric/symbol responses:* Replace "YES" or "NO" answers with numeric or special symbol responses (e.g., 1 or $ for "YES" and 0 or ~ for "NO").

Principle 3: *Comprehensive entity labels setting:* list all possible labels of target entities rather than a subset in NER. For example, in CDR, even if only disease and chemical are used, it is recommended to list all entity labels occurred in the document.

Principle 4: *Positive scenario - synonym expansion:* Expand the keywords of the relation to include all possible synonyms, e.g., "marker" for the relation "Gene-Disease: Marker/mechanism" can be interpreted as "mutate correlate," "mediate progression", etc.

Principle 5: *Positive scenario - verbal form variation:* Expand the key verbs of the relation into multiple different expressions, including passive forms and noun forms. For example, for the relation "Chemical-Disease: induce", the following variations can be used: "Disease is induced by Chemical," or "Chemical is an induction of Disease".

Principle 6: *Negative scenario - Error entity substitution:* Replace the {head} and {tail} entities with any possible error entities to create negative scenarios. a positive scenario might be "{gene} is effective in treating {disease}," but LLM might incorrectly treat "glutathione (a chemical) is effective in treating hepatotoxicity through its interaction with Nrf2" as a positive scenario. To avoid this, add a negative scenario such as "a chemical is effective in treating {disease}".

Principle 7: *Negative scenario - Antonym substitution:* Replace keywords with their antonyms to create negative scenarios, e.g., a positive scenario might be "{chemical} enhances {gene} expression" but an LLM might incorrectly treat "{chemical} reduces {gene} expression]" as a positive scenario. Therefore, add "{chemical} reduces/ downregulates {gene} expression]" as a negative scenario.

Principle 8: *Negative scenario - head-tail exchange:* Swap the positions of the {head} and {tail} entities to create negative scenarios, as LLMs might mistakenly treat "{chemical} affects {gene}" and "{gene} affects {chemical}" as the same meaning.

Principle 9: *Negative scenario - confusing term substitution:* Replace biomedical terminology with terms that may cause confusion, e.g., LLMs might confuse "gene expression" with "gene activity," "gene transport," or "gene excretion". Therefore, create negative scenarios using these potentially confusing terms.

In Section *RE Scenario Evaluation*, we introduce the evaluation methods for scenario assessment.

### *Datasets*

We evaluate our approach on two benchmark biomedical datasets: ChemDisGene(Zhang et al., 2022) is a biomedical multi-label document RE dataset. It consists of 76942 distantly supervised training set and 523 fully expert-labeled test set. We use its test set for evaluation the performance verse existing methods. CDR(Li et al., 2016) is a binary interaction dataset proposed by BioCreative-V to predict the relation between chemicals and diseases, consisting of 1500 biomedical documents. The training, development and test sets are 500 documents each. Table 3 shows the stats for both datasets. We only use test set for evaluation the performance verse existing methods.

**Table 3** Data Statistics

| Dataset Name | Annotation method | #Training set | # Test set | Relations |
|---|---|---|---|---|
| ChemDisGene | distantly supervised for training set, human annotated for test set. | 76942 | 523 | 14 |
| CDR | human annotated | 1500 | 500 | 1 |

### *Implementation*

Due to information security needs, especially in the biomedical field, we only consider localizable open LLMs. Llama is open for commercial and research use and able to be deployed in local environment, therefor we utilize Llama 3.1-70b-instruct(Grattafiori et al., 2024) deployed on Ollama for main experiments. We run our experiments on two Nvidia RTX A6000 GPU cards, costing approximately $13,000. Hyperparameter *Temperature* is set to zero.

## Results and Discussions

### *Results on ChemDisGene*

We compare with existing supervised models trained by distantly supervised for training set and evaluated by expert-labeled test set (curated corpus).

**Table 4** RE F1 Result on ChemDisGene Test Dataset

| Model | Micro | | | Macro | | |
|---|---|---|---|---|---|---|
| | P (%) | R(%) | F1(%) | P(%) | R(%) | F1(%) |
| BRAN* | 41.8 | 26.6 | 32.5 | 37.2 | 22.5 | 25.8 |
| PubmedBert* | 64.3 | 31.3 | 42.1 | 53.7 | 32.0 | 37.0 |
| PubmedBert+ BRAN* | 70.9 | 31.6 | 43.8 | **69.8** | 32.5 | 40.5 |
| ATLOP* | **76.17±0.54** | 29.7±0.36 | 42.73±0.36 | - | - | - |
| SSR-PU* | 54.27±0.40 | 43.93±0.32 | 48.56±0.23 | - | - | - |
| TTM-RE* | 53.83±0.85 | 53.34±0.15 | 53.59±0.27 | - | - | - |
| Ours | 53.12±0.07 | **62.48±0.19** | **57.42±0.07** | 50.25±0.15 | **56.41±0.23** | **52.18±0.13** |

Note: P represents Precision, R represents Recall.
- indicates unpublished data, ∗ are taken from TTM-RE accordingly

Table 4 shows that F1 of our approach is higher than exist methods in both micro and macro results. Recall of exist methods is lower than precision caused by a large number of false negatives for existing models are trained by the incomplete labeling phenomenon in the training set. However, it does not impact our approach. In the contrast, recall of our approach is much higher (+9%) than TTM-RE(Gao et al., 2024) and other exist methods.

Zhang et al.(Zhang et al., 2022) observed macro results are lower than Micro in Table 4, indicating that performance varies across different relation types, particularly low frequencies in the training data tend to perform poorly, e.g. bad performance of PubmedBert+BRAN on Chemical-Gene: expression-affects (0.4%). However, this issue does not impact our approach.

Fig 4 demonstrates that F1 of this relation in our approach is 48.6%, indicating that performance does not correlate with data frequencies.

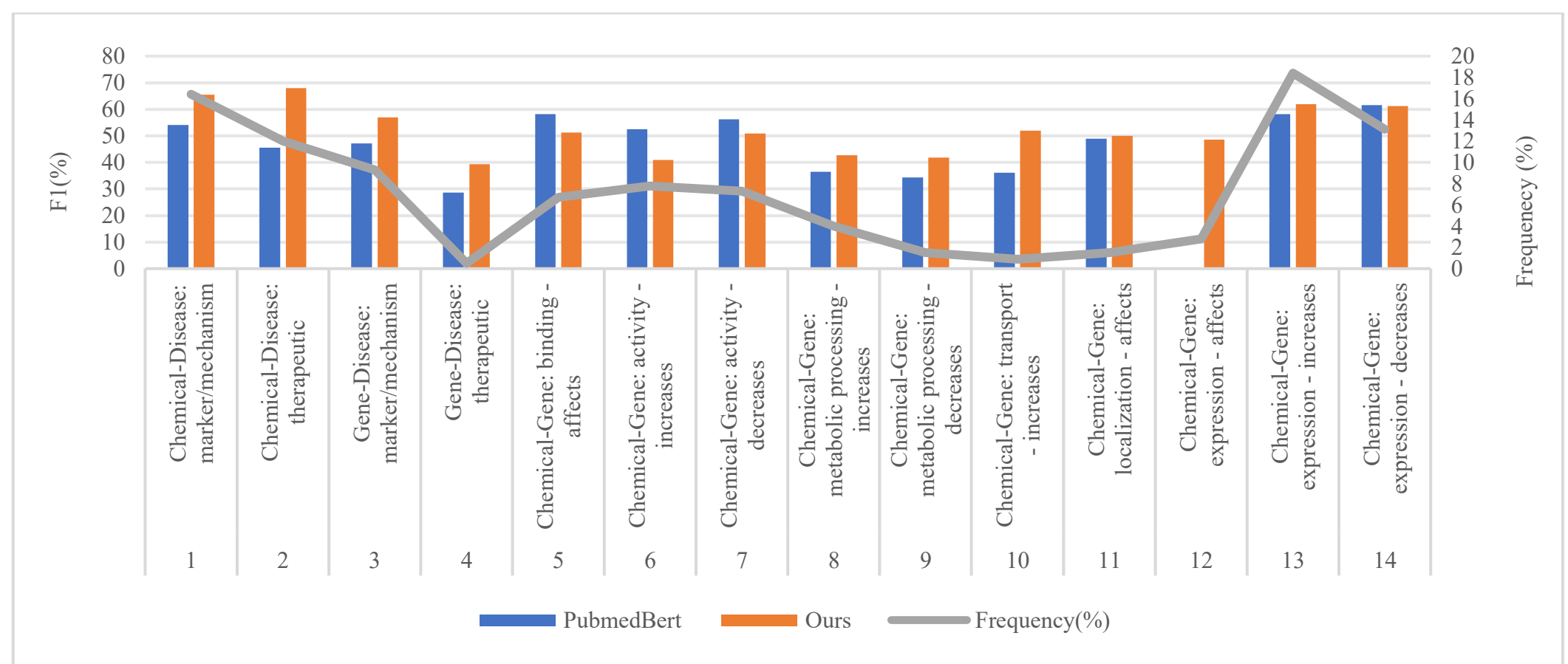

**Fig. 4** F1 comparison with PubmedBert+BRAN on each relation

However, we identified two factors that impact our results: Diversity of Language Expression: The diversity of language expression for a relation impacts recall. For instance, biomarker can be expressed in various ways: increased expression of gene X correlates with breast cancer, mutations in gene X cause liver cancer, etc. This diversity in expression leads to an increase in false negatives when not all scenarios are covered. LLM Error Response: Errors in the responses of LLMs impact precision. For example, LLM may mistakenly regard a chemical or another gene treating disease as a positive scenario for gene treat disease. To mitigate this.

We summarized these scenarios into scenario-based prompt design principles.

### *Results on CDR*

Fig. 5 shows RE prompt template for CDR. We designed 2 positive scenario and 2 negative scenarios.

{input}

You are biomedical expert, refer to above biomedical article, please give answer '$' or '~' only based on below conditions:

**Assumptions:**

1.1 if 'patient had/exhibited {tail} on the Xth day of {head} treatment', please do not use it and find other reason.

1.2 if 'difference did not reach statistical significance compared to groups' appeared, please do not use it and find other reason.

**Conditions:**

**Positive Scenario 1:** if the article gave the conclusion that the interaction between {head} and another chemical induce {tail}, or {head} induce {tail}, answer '$'.

**Positive Scenario 2:** else if the article mentioned {tail} was reported in the percentage of patients because of {head}, answer '$'.

**Negative Scenario 1:** else the article mentioned '{head} is mediating/attenuating {tail}, answer '~'

**Negative Scenario 2:** else if {head} and {tail} have no relationship, answer '~'.

DO NOT output any other message.

**Fig. 5** RE prompt template for CDR

**Table 5** RE F1 Results on CDR Test Dataset

| Model | P (%) | R(%) | F1(%) |
|---|---|---|---|
| Graph-based | | | |
| EoG* | 62.10 | 65.20 | 63.60 |
| DHG* | 61.80 | 70.50 | 65.90 |
| GLRE* | 65.30 | 72.30 | 68.60 |
| Knowledge-based | | | |
| RC* | 71.93 | 70.45 | 71.18 |
| KCN* | 69.70 | 73.00 | 71.30 |
| KGAGN* | 71.80 | 75.00 | 73.30 |
| Transformer-based | | | |
| MGSN+SciBERT * | 67.70 | 69.70 | 68.70 |
| ATLOP+SciBERT * | 68.30 | 70.10 | 69.20 |
| DocuNet+SciBERT * | 77.41 | 75.23 | 76.31 |
| DocuNet+BioBERT* | 81.79 | 73.36 | 77.34 |
| MMR+SciBERT * | **82.71** | 74.95 | 78.64 |
| MMR+BioBERT* | 81.52 | 76.55 | **78.95** |
| LLM-based | | | |
| Ours | 75.07 | **81.19** | 78.01 |

Note: ∗ are taken from MMR accordingly.

Our results on the CDR dataset are shown in Table 5. The existing supervised model such as MMR(Zhang et al., 2024) model has a SOTA result. Notably, our F1 is very close to that of MMR (-0.63%), and we achieve a higher recall than MMR(+4.64%) . This means our method can identify more induced-disease chemicals. In the biomedical field, recall rate is sometimes more crucial than precisions.

### *Ablation study*

#### Synonym and Hypernym Evaluation

Table 6 presents an ablation study evaluating the impact of synonym and hypernym models on relation extraction performance using the CDR dataset.

**Table 6** Ablation Study for Synonyms and Hypernyms on CDR

| NER Model | P(%) | R(%) | F1(%) |
|---|---|---|---|
| w/ synonym + hypernym | 75.07 | 81.19 | 78.01 |
| w/o synonym | 54.66 | 80.89 | 65.23 |
| w/o hypernym | 57.18 | 79.69 | 66.58 |

The complete model incorporating both synonym and hypernym achieves a precision of 75.07%, recall of 81.19%, and F1 score of 78.01%. When the synonym model is removed, precision drops significantly to 54.66%, while recall decreases slightly to 80.89%, resulting in an F1 score of 65.23%. This indicates that the synonym model plays a crucial role in enhancing precision, likely by accurately identifying and normalizing synonymous entity mentions, thereby reducing false positives.

Similarly, removing the hypernym model leads to a precision of 57.18%, recall of 79.69%, and F1 score of 66.58%. The decline in performance suggests that the hypernym model contributes to both precision and recall, possibly by leveraging hierarchical semantic relations to better disambiguate entity relations and capture more comprehensive relational contexts.

#### Entity labels setting for NER

In our observations, while it is true that only chemical and disease entities are required for the CDR, restricting the NER task to solely these two labels led to a significant number of false negatives. For instance, DNA has been erroneously categorized as a chemical. To address this issue, we have expanded the label set by incorporating additional categories such as treat, physiology, immunity, and gene. The detailed outcomes of this modification are presented in Table 7.

**Table 7** Evaluation Result with Different NER Entity Labels Settings on CDR

| Entity label | Disease | | | Chemical | | |
|---|---|---|---|---|---|---|
| | P(%) | R(%) | F1(%) | P(%) | R(%) | F1(%) |
| disease, chemical, treat, physiology, immune, gene | 90.39 | 77.62 | 84.52 | 91.52 | 76.24 | 83.18 |
| disease, chemical | 88.73 | 76.97 | 82.43 | 88.23 | 76.70 | 82.06 |

LLM utilizes normalized function like Softmax to transform an original vector with arbitrary real values into a probability vector with real values in the range [0, 1] that add up to 1. Softmax function is defined as:

$$P(C_i) = \frac{e^{z_i}}{\sum_j e^{z_j}} \quad (4)$$

Given a label $C_i$, the entity has a possibility $P$ based on the logit value $z_i$. So even the words do not belong to any label, it might still be labeled based on max of $P$. To avoid this error, it is necessary to list all possible labels happened in the document.

**Two Stage method vs One Stage method**

To evaluate the efficacy of our Two Stage approach, we conducted an experiment using a One Stage method. This method merge Stage I and Stage II into a single prompt which aims to extract relations from end to end. The prompt template is depicted in Fig. 6.

| {input}<br>You are biomedical expert, refer to above biomedical article, please follow below step to give the final output.\ndefine entity as object with name, type ='Disease' or 'Chemical', index default =0, flag default ='-' |
|---|
| **STEP 1. Extract Entity: extract all entities from above context by below types:** |
| Positve Scenario 1: Disease: Disease/Abnormal/Symptom/Adjectives/abbreviations for diseases such as {Disease_include}. |
| Postive Scenario 2: When an organ name ahead of disease name, please include the organ. |
| Negative Scenario 1: Exclude {Disease_skip} |
| Postive Scenario 3: Chemical:Chemicals/Drugs/antigen/abbreviations names, such as {Chemical_include} |
| Negative Scenario 4: exclude {Chemical_skip}. |
| Attention: the entity must be appeared in the text. please do not output. |
| **STEP 2. Synonym Recognition: Review all entities,** |
| Postive Scenario 1: If one entity is plural form of another, |
| Postive Scenario 2: else if one entity is the alias name of another, |
| Postive Scenario 3: else if one entity is a abbreviations of another, |
| please give same index to both entities, otherwise please give different index to the entity.please do not output. |
| STEP 3. Hypernym Recognition: |
| 3.1 if one disease entity is a symptom of another disease entity, |
| 3.2. else if one disease is a hypernym of another disease in term of MeSH, assign the entity flag='H' If one chemical entity is one of a agonist/a receptor/a drug/a derivative of another chemical entity,please assign the entity flag='H' .please do not output. |
| **STEP 4. Relationship Extract based on below conditions:** |
| Conditions: |
| Postive Scenario 1. if the article gave the conclusion that the interaction between one chemical entity and another chemical induce one disease entity, or chemical entity induce one disease entity, |
| Postive Scenario 2. else if the article mentioned one disease entity was reported in the percentage of patients because of one chemical entity , |
| please output \| chemical entity name \| chemical entity index \| chemical entity flag \| 'induce' \| disease entity name \| disease entity index \| disease entity flag \| |
| Negative Scenario 1: else the article mentioned 'one chemical entity is mediating/attenuating one disease entity or no relationship, please ignore it. |
| please only output STEP4 result. |

**Fig. 6** Single prompt template from end to end

Table 8 shows that Two Stage method demonstrates a significant performance advantage over One Stage method.

**Table 8** Results of Two Stage Method vs One Stage Method

| **Methods** | **CDR** | | |
|---|---|---|---|
| | **P(%)** | **R(%)** | **F1(%)** |
| Two Stage method | 75.07 | 81.19 | 78.01 |
| One Stage method | 46.17 | 55.23 | 50.29 |

**RE Scenario Evaluation**

We design both positive scenario and negative scenarios based on scenario-based prompt design principles and evaluate their impact on performance metrics. Take the relation "Gene-Disease: therapeutic" as an example, we identified all positive scenarios and negative scenarios to form comprehensive scenarios set. We then evaluate precision, recall and F1 as baseline metrics ($P_{\text{baseline}}$, $R_{\text{baseline}}$, $F1_{\text{baseline}}$). Subsequently, we remove the $n^{\text{th}}$ scenario and re-evaluate the metrics ($P_n$, $R_n$, $F1_n$) to quantify the impact of each scenario. The evaluation function is defined as:

$$f_{evaluation} = \begin{cases} \Delta_{Precision} = P_{baseline} - P_n \\ \Delta_{Recall} = R_{baseline} - R_n \\ \Delta_{F1} = F1_{baseline} - F1_n \end{cases} \quad (5)$$

By iteratively removing scenarios with $\Delta_{F1} \leq 0$, we refine the scenario set to maximize the F1 score. This process helps identify and retain only the most effective scenarios that contribute positively to the model's performance.

Fig. 7 illustrates how positive and negative scenarios impacted the evaluation metrics. For example, positive scenarios (e.g., P1 and P2) with a large $\Delta_{F1}$ significantly boosts recall while slightly reducing precision. Removing this scenario would lead to a substantial drop in recall, indicating its crucial role in capturing true positive instances. In contrast, negative scenarios (e.g., N1-N12) primarily improved precision, removing them would result in more false positives. The bubble size represents $\Delta_{F1}$, with larger bubbles indicating a greater impact on the F1. The larger bubbles associated with P1 and P2 show that their removal would lead to a more significant decrease in F1, highlighting their importance in maintaining overall performance.

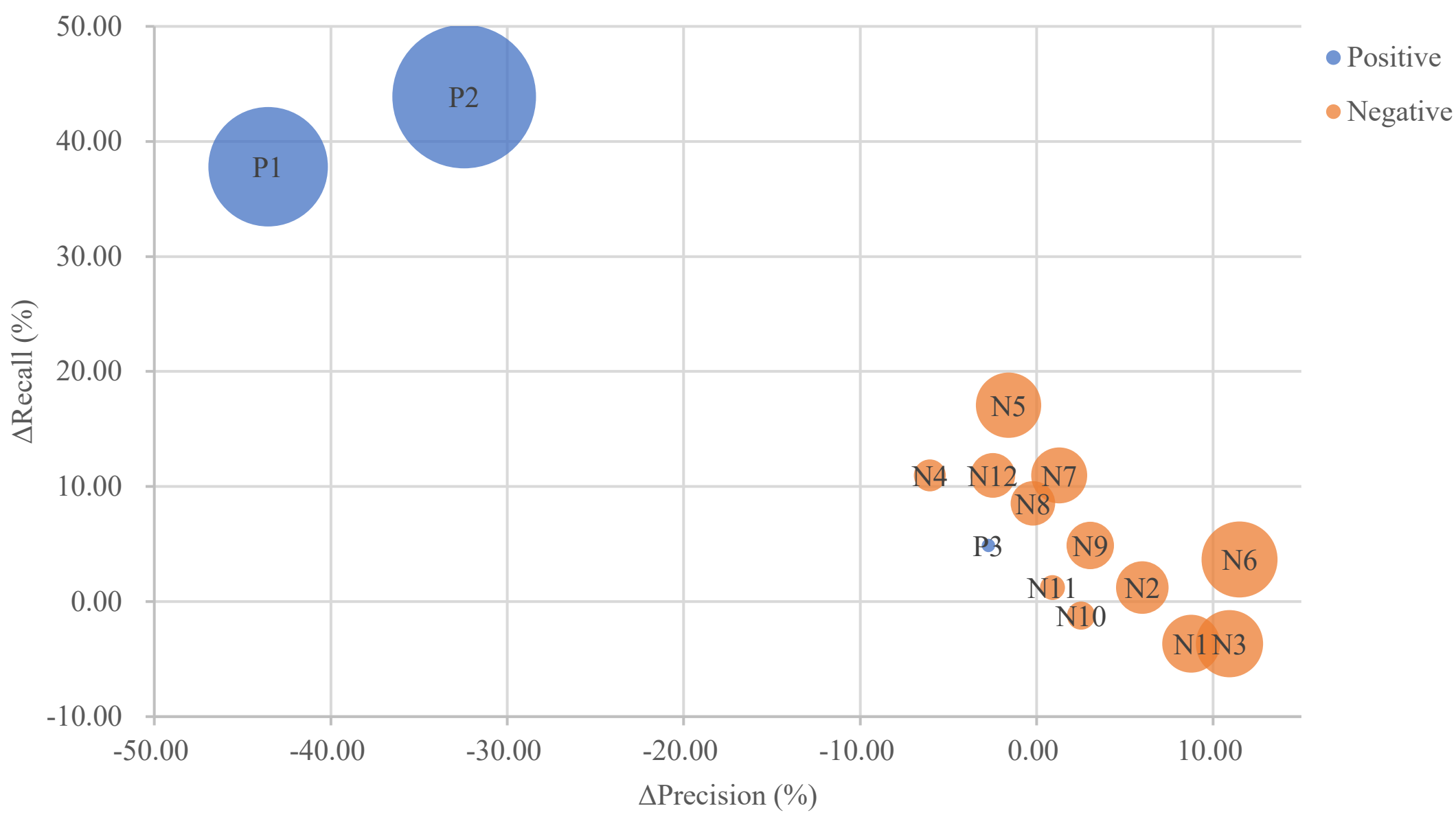

**Fig. 7** Impact of Removing Scenarios on Gene-Disease: therapeutic

### *Reducing Parameter Size and Evaluating LLM for RE*

We endeavored to reduce the parameter size from 70 billion to 32 billion, 27 billion, 14 billion to assess the viability of cost reduction. Additionally, we included Qwen 2.5 (72 billion parameters), which is similar in size of Llama 3.1, to demonstrate our approach is not limited to single LLM.

LLM selection strategy: we selected only open localizable LLM and exclude closed-source LLM such as OpenAI GPT series, Google Gemini.

**Table 9** RE Task Results with Different Parameter Size of LLMs

| LLM | Size | ChemDisGene | | | CDR | | |
|---|---|---|---|---|---|---|---|
| | | P(%) | R(%) | F1(%) | P(%) | R(%) | F1(%) |
| Qwen 2.5 | 72b | 47.49 | 54.43 | 50.72 | 76.74 | 77.36 | 77.05 |
| Llama 3.1 | 70b | 53.12 | 62.48 | 57.42 | 75.07 | 81.19 | 78.01 |
| Qwen 2.5 | 32b | 40.52 | 53.50 | 46.11 | 76.89 | 73.64 | 75.23 |
| Gemma2 | 27b | 41.84 | 2.64 | 4.98 | 65.71 | 87.12 | 74.91 |
| Phi-4 | 14b | 21.48 | 7.2 | 10.81 | 72.11 | 70.22 | 71.15 |

Key insights from Table 9:

Parameter Size and Performance: Larger models generally exhibit better performance in terms of precision, recall, and F1 scores, especially on the more complex ChemDisGene dataset.

Efficiency vs. Effectiveness: Smaller models (14B-32B parameters) offer a trade-off between performance and computational efficiency. While their performance is lower on ChemDisGene, they achieve reasonable results on CDR with substantially reduced processing time. As more advanced and larger LLMs continue to be released, we anticipate that even more accurate results will be attainable, further pushing the boundaries of what is possible in automated knowledge extraction from complex biomedical documents.

### Future work

The current reliance on manual processes for prompt design and iterative evaluation incurs substantial computational overhead, particularly when handling complex biomedical relationship schemas such as the relation between chemical and gene. Looking ahead, our future work will focus on refining and automating the entire relation extraction pipeline. This includes developing sophisticated prompt generation mechanisms based on our established scenario-based prompt design principles, as well as creating robust evaluation frameworks to systematically assess and filter scenarios. By optimizing these components, we aim to maximize F1 scores and achieve optimal performance with minimal manual intervention, ultimately making the relation extraction process more efficient and scalable for real-world applications in the biomedical field.

## Conclusion

In summary, this paper introduces a comprehensive framework for zero-shot document-level biomedical relation extraction using LLM. Our contributions include a two-stage framework, a five-part prompt template structure and a scenario-based prompt design principle. Through comprehensive evaluations and ablation studies, we have demonstrated that localized LLMs with substantial parameters can achieve competitive performance in document-level biomedical relation extraction tasks. Our findings highlight the effectiveness of our approach in leveraging the power of LLMs while addressing practical constraints such as computational efficiency and information security and offering a cost-effective solution.

## Data Availability

Data is available on https://github.com/tyrone1979/LLMRE.

## Acknowledgments

We are grateful for the support from the Neusoft Research Institute of Dalian Neusoft University of Information.

## Author Contributions

LZ and LK designed the methodology. QG supervised the project. LZ performed the formal analysis and initial data interpretation. LZ and LK conducted additional data analyses and contributed to software deployment. QG supervised the data curation, provided critical resources, and validated the findings. LZ and LK drafted the manuscript. QG reviewed and edited it. All authors approved the final manuscript before submission.

## Funding

This work was supported by the Dalian Science and Technology Innovation Fund Program [2022JJ12GX017], the United Foundation for Medicoengineering Cooperation from Dalian Neusoft University of Information and the Second Hospital of Dalian Medical University [LH-JSRZ-202201] and Technology Innovation Project of Dalian Neusoft University of Information [TIFP202302].